\documentclass[conference]{IEEEtran}

\usepackage{cite}
\usepackage{amsmath,amssymb,amsfonts}
\usepackage{algorithmic}
\usepackage{graphicx}
\usepackage{textcomp}
\usepackage{xcolor}
\usepackage{tabularx}
\usepackage{caption}
\def\BibTeX{{\rm B\kern-.05em{\sc i\kern-.025em b}\kern-.08em
    T\kern-.1667em\lower.7ex\hbox{E}\kern-.125emX}}
\begin{document}

\title{Finding the Perfect Fit: 
Applying Regression Models to ClimateBench v1.0\\
{\footnotesize \textsuperscript{}}
\thanks{}
}

\author{\IEEEauthorblockN{Anmol Chaure}
\IEEEauthorblockA{\textit{Research Scholar, CSE Department} \\
\textit{BIT Durg, India}\\
anmolchaure@bitdurg.ac.in}
\and
\IEEEauthorblockN{Ashok Kumar Behera}
\IEEEauthorblockA{\textit{Associate Professor, CSE Department} \\
\textit{BIT Durg, India}\\
ashok.behera@bitdurg.ac.in}
\and
\IEEEauthorblockN{Sudip Bhattacharya}
\IEEEauthorblockA{\textit{Associate Professor, CSE Department} \\
\textit{BIT Durg, India}\\
sudip.bhattacharya@bitdurg.ac.in}
}

\maketitle

\begin{abstract}

Climate projections using data driven machine learning models acting as emulators,  is one of the prevailing areas of research to enable policy makers make informed decisions. Use of machine learning emulators as surrogates for computationally heavy GCM simulators reduces time and carbon footprints. In this direction, ClimateBench\cite{ArticleAuthor2022ETHCitation} is a recently curated benchmarking dataset for evaluating the performance of machine learning emulators designed for climate data. Recent studies have reported that despite being considered fundamental, regression models offer several advantages pertaining to climate emulations. In particular, by leveraging the kernel trick, regression models can capture complex relationships and improve their predictive capabilities. This study focuses on evaluating non-linear regression models using the aforementioned dataset. Specifically we compare the emulation capabilities of three non linear regression models. Among them, Gaussian Process Regressor demonstrates the best in class performance against standard evaluation metrics used for climate field emulation studies. However, Gaussian Process Regression suffers from being computational resource hungry in terms of space and time complexity. Alternatively, Support Vector and Kernel Ridge models also deliver competitive results and but there are certain trade-offs to be addressed. Additionally, we are actively investigating the performance of composite kernels and techniques such as variational inference to further enhance the performance of the regression models and effectively model complex non-linear patterns, including phenomena like precipitation.

\end{abstract}


\section{Introduction}

The depletion of the ozone layer sparked an intense scientific investigation into viewing the Earth as a comprehensive system. The first extensive evaluation of global climate change was documented in the Charney Report\cite{J.G.Charneyetal.1979CarbonAssessment}, based on the circulation model developed by Syukuro Manabe\cite{Manabe1975TheModel}. This report concluded that the doubling of CO\(_2\) concentration would lead to a 3\textdegree C rise in the global temperature.
By the 1980s, it was evident from the Keeling curve\cite{Thoning1989Atmospheric1974-1985} that the level of carbon dioxide(CO\(_2\)) was increasing,pointing directly to the impact of anthropogenic activities.

The fundamental physical equations that governed the  Manabe circulation system have remained constant\cite{Rolnick2019TacklingLearning}.However, with the advancement of digital computers, significant progress has been made in terms of refining and enhancing climate predictions. General Circulation Models(GCMs) and Earth system models(ESMs) have been the forefront models for climate projection for several decades.These models have allowed decision makers to explore the Shared Socioeconomic Pathways(developed by IPCC in accordance with the Paris Climate Agreement), for the mitigation(Controlling emissions) and adaptations(Being equipped for the unavoidable effects) of Climate Change\cite{Rolnick2019TacklingLearning,Lenton2016EarthIntroduction}.

General Circulation Models (GCMs) have been criticized for their computational expensiveness and iterative calibrations.Furthermore, the models are unable to consider the climate fluctuations from the past data which may hold important lessons, leading to an enhanced role for data-driven(Machine Learning) emulators.

Machine learning emulators are able to overcome these challenges, by their ability to train and run faster, better sensor calibration, and increased accuracy and pattern recognition due to the availability of petabytes of data generated using the Coupled Model Intercomparison Project (CMIP6).
Regression Models , although simple, are still very popular machine learning models  and have demonstrated good performance across many climate datasets. In this work we compare three non linear regression models: Support Vector Regression, Kernel Ridge Regression and Gaussian Process Regression on  ClimateBench v1.0. These  models were selected for this study because of  their ability to train efficiently on non-linear and sparse samples. Furthermore, Gaussian Process Regression\cite{BagnellGaussianProcesses} exhibits the ability to quantify uncertainty, which is an essential feature for climate datasets.

\section{Climate Change}
In 1986, Svante Arrhenius\cite{Arrhenius1896OnGround} developed a climate model revealing the relationship between atmospheric CO\(_2\) and temperature change. His model demonstrated that atmospheric CO\(_2\) increases geometrically, which in turn leads to a nearly arithmetic progression of temperature change. Arrhenius projected that the doubling of CO\(_2\) solely from fossil fuel combustion would occur within a span of 500 years and result in a global warming of approximately 5°C.
There has been an exponential rise in global warming since. The evidence of climate change have been increasing, storms, forest fires, and floods are becoming stronger and more frequent. Climate Action has been designated as the $13^{th}$ Sustainable Development Goal by the United Nations, recognizing the immense challenges faced by the most vulnerable populations due to the impacts of climate change. Governing bodies like IEA(International Energy Agency) and IPCC(Intergovernmental Panel on Climate Change)\cite{IntergovernmentalPanelonClimateChangeIPCC2023ClimateVulnerability} call for urgent action.The IEA has recently devised an ambitious net-zero emission pathway\cite{EnergyAgency2050NetSector} that guides the transition toward achieving a carbon-neutral future.\cite{MaslinClimateIntroduction}

\section{General Circulation Model(GCMs)}

General Circulation models, three-dimensional models that simulate and analyze the complex interactions and dynamics of the Earth's atmosphere and oceans, have been the forefront of climate modelling since their initial introduction by Syukuro Manabe and Kirk Bryan\cite{Manabe1975TheModel}. GCMs is utilizes the equations of fluid dynamics and thermal exchange to describe atmosphere-ocean interaction. 

\begin{equation}
    \frac{\partial x}{\partial t} = R(x) + U(x) + P(x) + F 
\end{equation}

where x is the state vector, R(x) + U(x) represents the Navier-Stokes equation, P represents the thermodynamic processes and F represents external forcings over the system.\cite{Balaji2022AreObsolete}
GCMs are employed to examine the climate system's response to variations in the parameter F. However, they have high computational cost, requires continuous tuning of hundreds of parameters, and fails to recognize the overall uncertainty of the climate pattern. \cite{Balaji2022AreObsolete, Kay2015TheVariability, Carman2017PositionCapability}

\section{Machine learning Emulators (surrogates)}

With the rise in availability of climate observation data produced by numerous satellites and climate modelling projects, Data-driven emulators are becoming the forefront for projections.
Machine learning emulators are used to model and approximate the behavior of complex systems. They address the challenges prevalent in GCMs, by using automatic calibration techniques, being computationally efficient and using new observations to enhance the accuracy of the models.

Regression is a widely recognized applied learning model. Any regressor aims to model the relationship between the regressor or independent variable(x) and the response or dependent variable(y). If there exists a relationship it is represented using the function f,

\begin{equation}
    y = f(x,\beta) + \epsilon
\end{equation}

Here, $\epsilon$ represents the error term or the residual(the differences between our predictions and real samples.), which accounts for the variability or randomness in the observed dependent variables that cannot be explained by the nonlinear function f and the parameters $\beta$.\cite{Seber1989NonlinearRegression}

In this paper, we use three non linear regression models, namely, Gaussian Process Regression, Support Vector Regression and Kernel Ridge Regression. Along with the ability to capture non linearity these models employ special mathematical function called "kernel". Kernels transform the original data into a new feature space where the complex and intricate patterns of data become more evident. This feature is especially useful in our use case of climate data.
Although the equation remains the same, but the kernel transformation helps us capture more nuanced and non-linear connections between the variables.

\section{Dataset}

Here, we have used ClimateBench, a recently developed benchmark specifically curated for evaluation of machine learning model with the primary objective to predict air surface temperature, diurnal temperature, precipitation and the 90th percentile precipitation. This prediction is done from the four most common forcings in the Anthropocene era, Carbon dioxide, Methane, Sulphur dioxide and Black Carbon. 

ClimateBench is devised using a selection of simulations like ScenarioMIP, CMIP6(Coupled Model Intercomparision Project, Phase-6) and DAMIP(Detection and Attribution Model Intercomparision Project).This dataset does not provide temporal predictions, but rather offers climate projections based on the scenarios designed by the IPCC.

\begin{figure}[h]
    \centering
    \includegraphics[scale = 0.30]{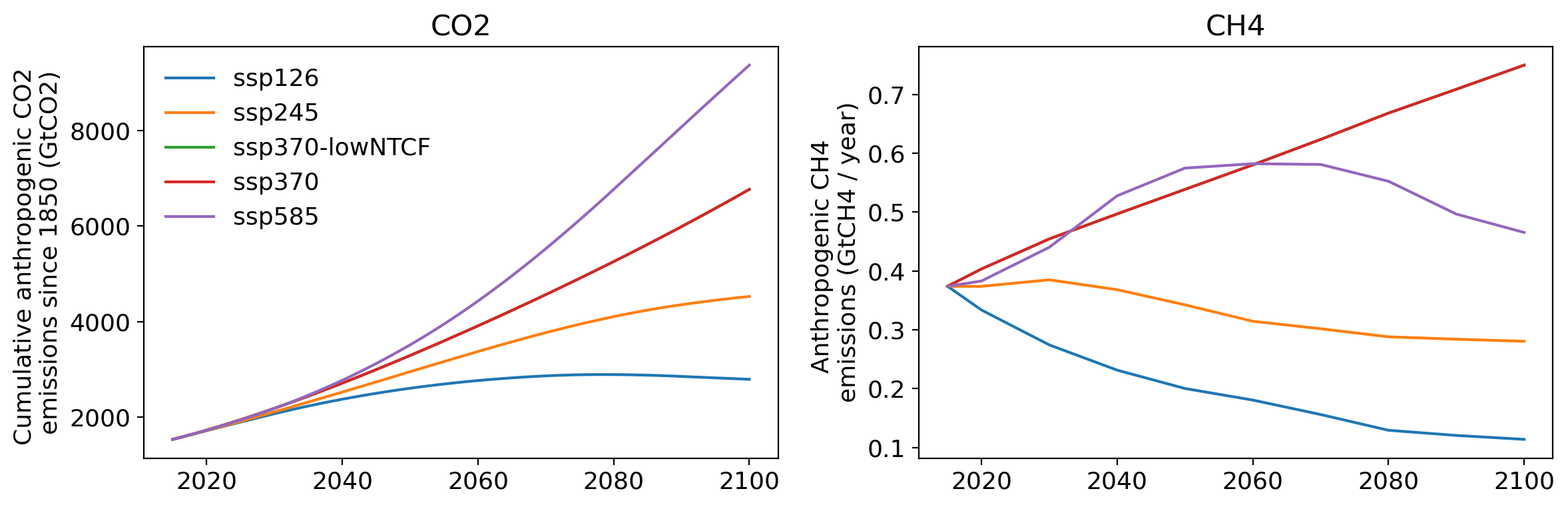}
    \includegraphics[scale = 0.30]{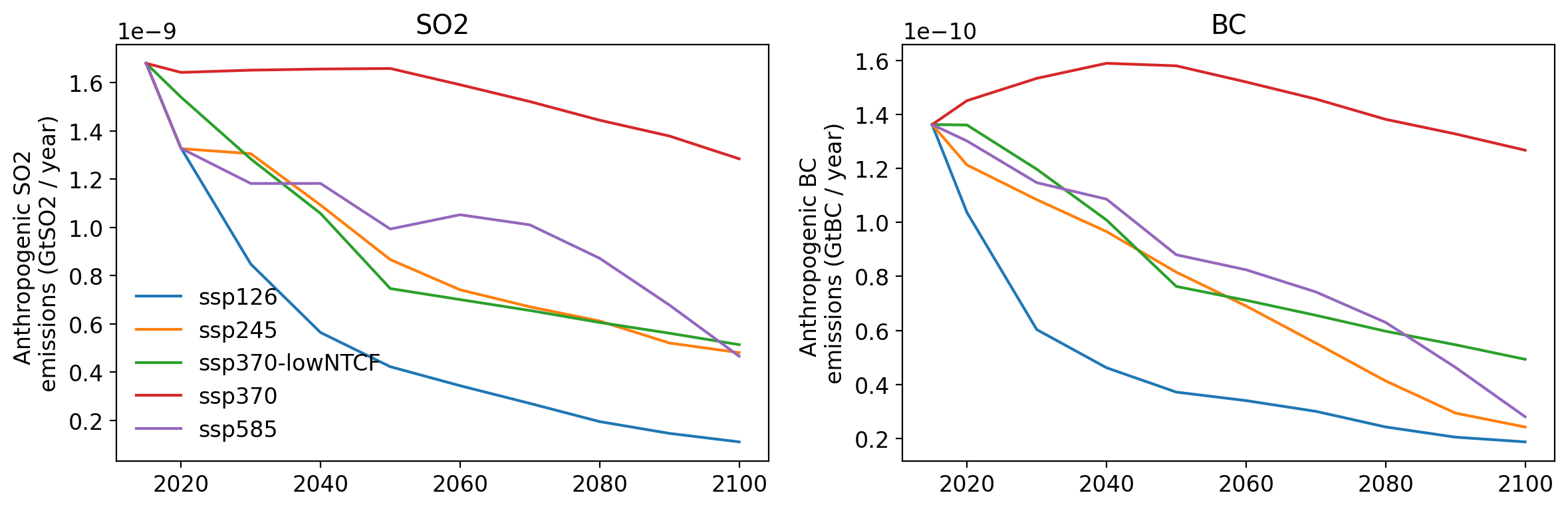}
    \caption{Emission Scenarios in ClimateBench}
    \label{eda}
\end{figure}

\section{Gaussian Process Regression}
Gaussian process regression(GPR)\cite{Rasmussen2006GaussianLearning} is a supervised learning, probabilistic framework.Similar to regression modelling, it seeks to find the relationship between input and output variables. But, in addition to providing the predicted value GPR also provides a confidence interval.

\subsection{Working of GPR}

Gaussian processes provide an elegant solution to the problem of fitting data by assigning a probability to each function among the potentially infinite number of functions that can fit a given set of training points.\cite{Rasmussen2006GaussianLearning}
GPRs have the property of being closed under conditioning and marginalization, i.e, the output distributions will also be Gaussian distribution. \cite{Gortler2019AProcesses}

First, we need to concatenate the training(X) and testing(Y) data, to get the joint probability distribution $P_{x,y}$, as we want to learn the underlying distribution of Y given X. Therefore, this joint distribution consists of all possible function values that we want to predict. To perform regression, we have to treat it like a Bayesian inference, so that we would have to update the current prior knowledge to posterior when new observations(X) are available. Therefore, we need the conditional probability $P_{Y|X}$, and as GPRs are closed under conditioning the given distribution is also Gaussian.

The next task is to calculate the mean $\mu$ and covariance matrix $\Sigma$. For Gaussian distributions it is assumed that $\mu = 0$, if $\mu \neq 0$ we assume it to be 0. The covariance matrix gives us  information about the distribution of the data and also the characteristic of the function that we are predicting. To calculate the matrix we require a function called covariance function or kernel (k), this function takes two inputs x,x' and outputs their measure of similarity.

\begin{equation}
    \Sigma = cov(X,X') = k(x,x')
\end{equation}

This equation is calculated for each test point to create a covariance matrix. Since the kernel describes the similarity between the values of our function, it controls the possible shape that a fitted function can adopt. Kernel essentially plays an important role in the accurate modelling of the relationship between X and Y.

Through conditioning, we find $P_{Y|X}$, from $P_{X,Y}$, this distribution will have the same distribution as the test points $N = |Y|$, this provides us with a multivariate Gaussian distribution for the whole test dataset.
Through marginalization, we ensure that the probability distribution of multiple variables depends only on one variable(x), by integrating all possible values of all other variables at the value x. Calculate the mean and standard deviation of the distribution at each $i^th$ test datapoint.

Using this we have the mean and standard deviation of every point in Y, the confidence interval can now be calculated using the standard deviation to a desired level.
As we move away from the training datapoints, the prediction uncertainty gradually increases, while it remains relatively low in regions near the training data.\cite{Gortler2019AProcesses,Bailey2019GaussianDummies}

\subsection{GPR Kernels}

The accuracy and efficiency of the Gaussian Process predominately depend upon the kernel functions. In addition to the wide variety of available kernels, they can also be combined to fit the data better. The most used kernel combination is addition and multiplication. There exist numerous types of kernels, the most commonly used include Linear, Exponential, Matern(1/2,3/2,5/2), Radial basis function or Squared Exponential.\cite{DuvenaudTheFunctions}
In this project, the Matern 3/2 kernel was chosen for its ability to yield the most accurate results.

\section{Support Vector Regression(SVR)}

Support Vector Regression\cite{DruckerSupportMachines} is considered a non-parametric technique due to its use of kernel functions.The model produced by support vector classification  depends only on a subset of the training data, because the cost function for building the model does not care about training points that lie beyond the margin. Analogously, the model produced by Support Vector Regression depends only on a subset of the training data, because the cost function ignores samples whose prediction is close to their target.\cite{Zhang2020SupportRegression}
As a regression model, SVR tries to find a function f(x) = wx +b that best fits the model. In regression our goal is to quantify our prediction,therefore, we aim to have the data points (observations) as close as possible to the hyperplane, in contrast to SVM for classification. 

\begin{equation}
    |y(x)-f(x)| \leq \epsilon + C
\end{equation}

Here, f(x) is the predicted function, y(x) is the actual function, $\epsilon$ is the width of of marginal plane, C is the distance between data points outside marginal plane.

SVR is similar to simple regression techniques such as Ordinary Least Square, but with one key difference. We introduce an epsilon range on both sides of the hyperplane, which makes the regression function less sensitive to errors. Ultimately, SVR in regression has a boundary, similar to SVM in classification, but the purpose of the boundary in regression is to reduce the sensitivity of the regression function to errors. In contrast, the boundary in classification is designed to create a clear distinction between classes in the future (which is why it's referred to as the safety margin).

The efficiency of SVR depends heavily on the selection of appropriate kernel, because kernel selection helps in mapping the data into higher dimensional without actually calculating the coordinates for that space. Therefore, reducing the computational need and increasing the accuracy of mapping non linear data points.

\section{Kernel Ridge Regression(KRR)}

Kernel ridge regression (KRR) combines ridge regression (linear least squares with l2-norm regularization) with the kernel trick.KRRs are regression models that can capture both linear and nonlinear relationships between predictor variables and outcomes. These models are considered non-parametric, meaning they do not rely on strict assumptions about the underlying data distribution. However, the performance of kernel ridge regression heavily depends on the selection of hyperparameters.The objective of kernel ridge regression is to minimize the sum of squared errors between the predicted values and the actual outcomes, while also minimizing the complexity of the model. The model can be represented by the equation:

\begin{equation}
    f(x) = \Sigma (\alpha_i * k(x_i, x)) + b
\end{equation}

where f(x) is the predicted outcome for a new input x,   $\alpha_i$ are the learned coefficients, $x_i$ represents the training inputs, $K(x_i, x)$ is the kernel function that measures the similarity between $x_i$ and x in the feature space, and b is a bias term. To address this sensitivity, Kernel Ridge Regression offers a solution by employing k-fold cross-validation on predefined grids of hyperparameter values. This technique allows for a systematic evaluation of various hyperparameter combinations. By performing cross-validation, the optimal hyperparameter values can be determined, leading to more reliable and accurate predictions in the regression model.

It thus learns a linear function in the space induced by the respective kernel and the data. For non-linear kernels, this corresponds to a non-linear function in the original space.

\section{Related Work}

Recently, a generalized machine learning model ClimaX\cite{Nguyen2023ClimaX:Climate} was introduced for weather and climate. It presents  flexible and generalized deep learning model for weather and climate science can be trained using heterogeneous datasets spanning different variables, spatio-temporal coverage, and physical groundings.

Another paper by L.A Mansfield et al. \cite{Mansfield2020PredictingLearning} ,involves training of ridge and Gaussian regression using short-term simulations and historical climate data. By integrating relevant environmental variables and extracting meaningful features from the short-term simulations, the models are designed to capture the underlying patterns and relationships that contribute to long-term climate change. The authors carefully curate and preprocess the input features to ensure the models' robustness and generalizability.

ESEm v1.0.0\cite{Watson-ParrisModelEmulator}, an openly accessible and scalable Earth System Emulator library developed by Duncan Watson-Parris , incorporates GP fitting using the open-source GPFlow, a machine learning library that leverages Graphical Processing Units (GPUs) to expedite GP training, as mentioned earlier. The paper demonstrates the utilization of ESEm to generate emulated responses by employing GP emulation of aerosol optical depth (AAOD) with a 'Bias + linear' kernel and showcases its remarkable capability in accurately reproducing the spatial characteristics of AAOD while minimizing errors.

Additionally, the use of GPR in climate emulation is discussed\cite{Lalchand2022KernelScience}, the paper also explores the utilization of Gaussian Process Regression (GPR) in climate emulation, specifically focusing on its application in modeling and predicting precipitation patterns in the Hindu Kush Karakoram Himalayan region.

GPR has been utilized as a baseline emulator in ClimateBench\cite{ArticleAuthor2022ETHCitation}.Gaussian process regression is used to develop an emulator for a computationally expensive climate model. The emulator is then used to generate a large ensemble of simulations, which enables the exploration of the model's parameter space in a more efficient and systematic manner.

\section{Results}

Among the three regression models (SVR, GPR, and KRR) considered, GPR demonstrated the best performance.The performance of the models was assessed based on the RMSE(Root Mean Squared Error).These findings highlight the effectiveness and suitability of GPR for the given regression task.

\begin{table}[h]
\centering
\begin{tabularx}{\linewidth}{|r|X|X|X|X|X|X|X|}

\hline
\multicolumn{7}{|c|}{RMSE(Root Mean Squared Error)} \\

\hline
Year & 2050 & 2100 & 2045-2055 & 2090-2100 &  2050-2100 & \textbf{20Y average} \\

\hline
GPR(Climatebench) & 0.33 & 0.393 & 0.349 & 0.421 & 0.373 & \textbf{0.246} \\
\hline

\hline
GPR & 0.308 & 0.350 & 0.353 & 0.372 & 0.38 & \textbf{0.184} \\
\hline

\hline
SVR & 0.299 & 0.487 & 0.366 & 0.529 & 0.462 & \textbf{0.402} \\
\hline
KRR & 0.337 & 0.351 & 0.356 & 0.382 & 0.379 & \textbf{0.227} \\
\hline

\end{tabularx}
\hspace{10pt}

\captionsetup[table]{labelsep=period}
\captionsetup{format=plain}
\caption{RMSE values for different Regressors against different lead times}
\label{tab:rmse}
\end{table}

For the visualization of results generated using cartopy and panoply, please refer to the Appendix.

\section{Conclusion}
Machine learning emulators have demonstrated the capability to achieve comparable accuracy to large-scale Earth System models while requiring significantly less computational resources. However, each model has its own strengths and weaknesses, although they generally excel in evaluation metrics and effectively reproduce the temperature response of NorESEM2 in a realistic future scenario. Among these models, GPR stands out as the top performer in evaluation metrics and benefits from its ability to quantify uncertainty. Nevertheless, training the model to understand non-linear responses like precipitation has posed challenges.

Despite the notable performance of regression models, they are often considered fundamental. Researchers are more leveraging more advanced models such as neural networks, transformers, and generative models for climate projections. These advanced models come with a cost of increased carbon footprints. The objective of this work was to disseminate the effectiveness of simple regression models for climate projections with the purview of carbon footprint and model complexity. Results clearly denote the effectiveness of these models and the identifies Gaussian Process Regression as the best in class model for the task over the considered dataset.

\bibliographystyle{unsrt}
\bibliography{references}

\vspace{12pt}

\onecolumn

\newpage

\begin{center}
    
    \textbf{*A. Appendix}
    
\end{center}

\begin{figure*}[h]
  \centering
  \textbf{Results with Gaussian Process Regression}
  \includegraphics[width=\textwidth]{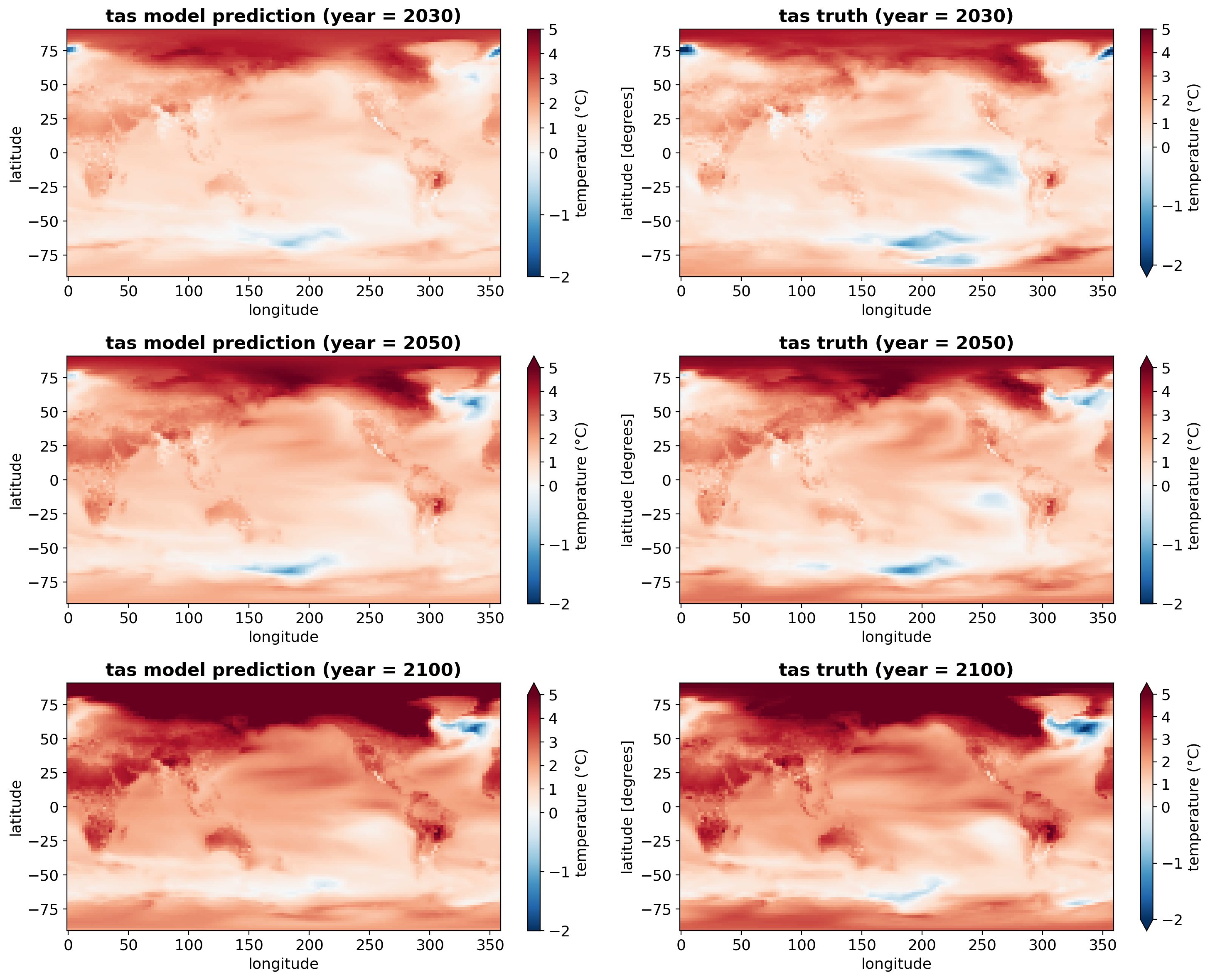}
 \caption{Results of Year Wise Prediction of Air Surface temperature using GPR for different lead times}
  \label{GPR_tas}

  \centering
  \includegraphics[width=\textwidth]{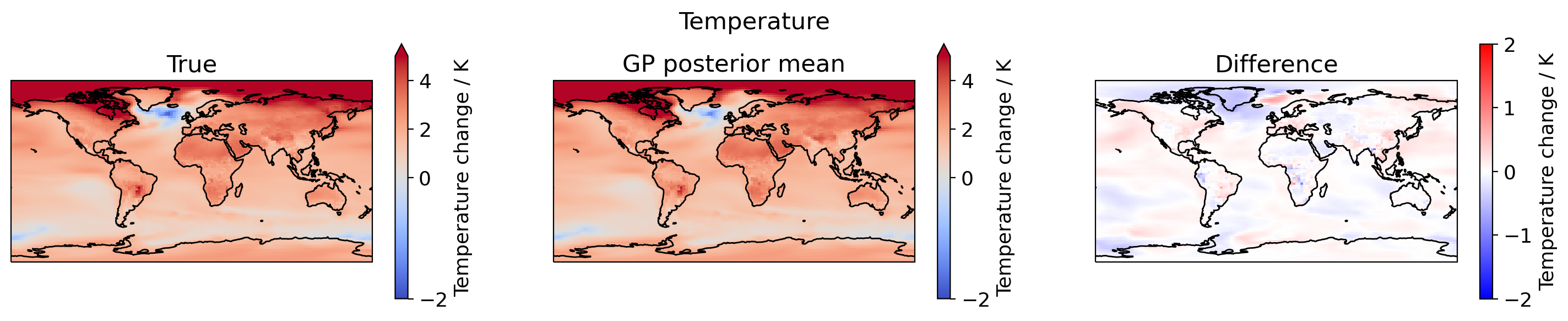}
  \caption{Difference of Air Surface temperature using GPR}
  \label{GPR_tas}
\end{figure*}

\begin{figure*}[h]
  \centering
  \textbf{Results with Support Vector Regression}
  \includegraphics[width=\textwidth]{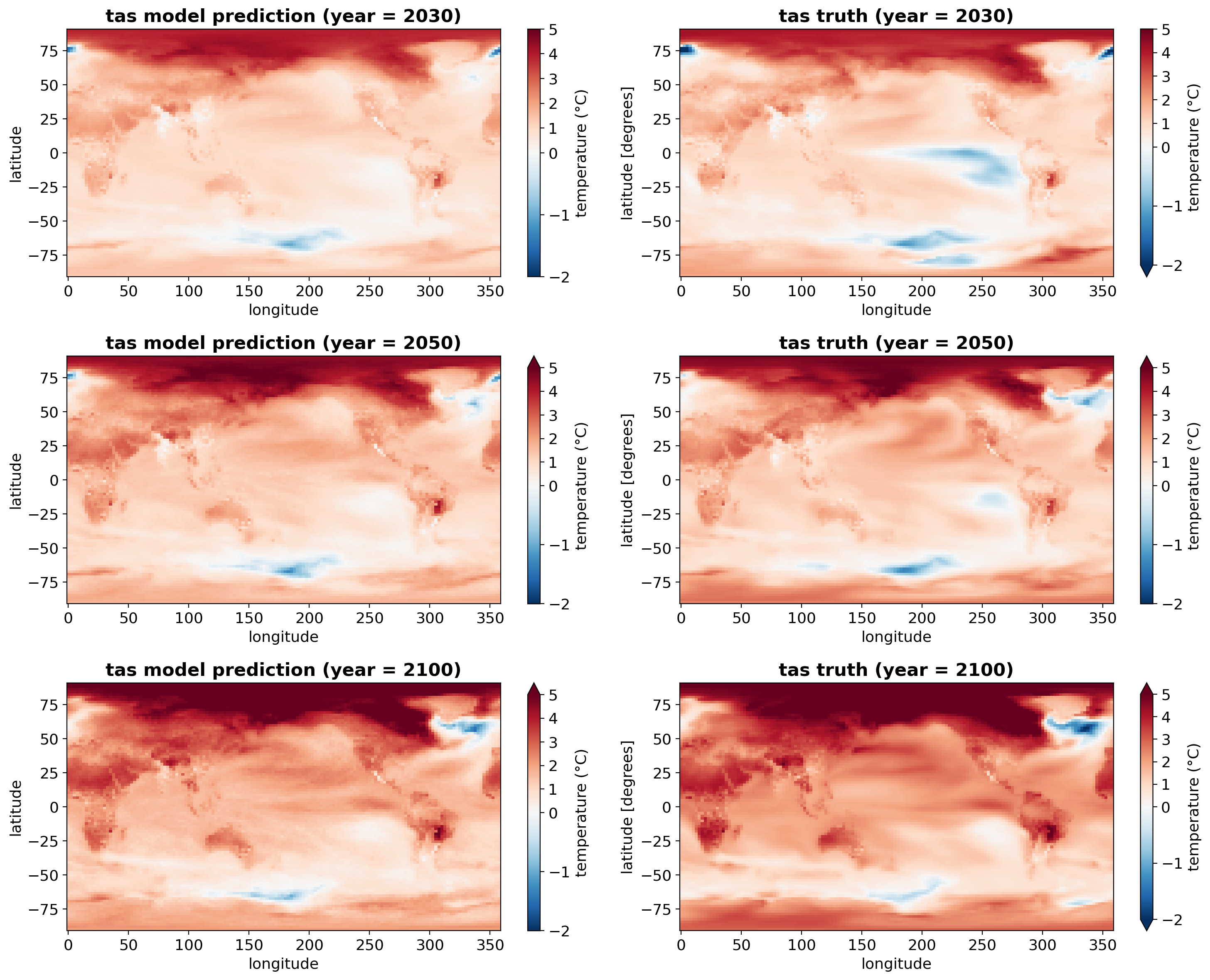}
  \caption{Results of Year Wise Prediction of Air Surface temperature using SVR for different lead times}
  \label{SVR_tas}
\end{figure*}

\begin{figure*}[h]
  \centering
  \includegraphics[width=\textwidth]{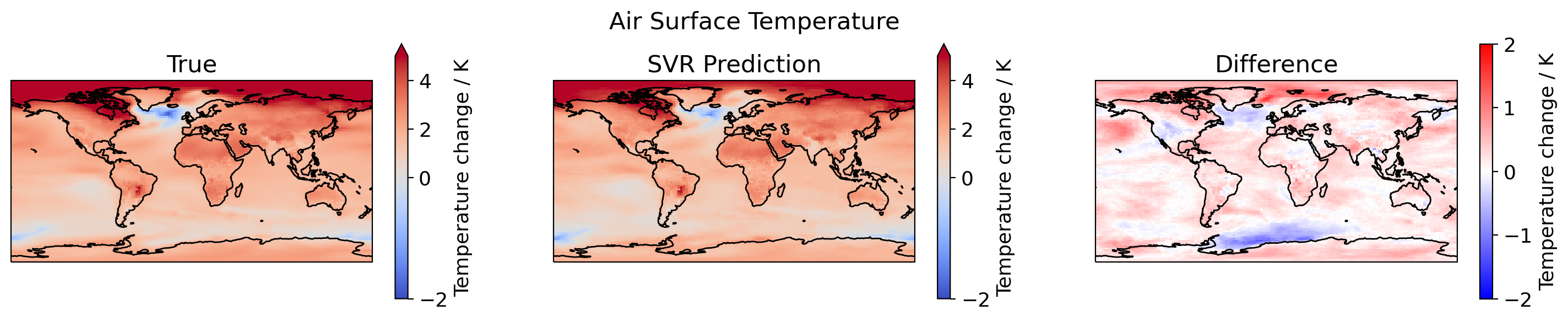}
  \caption{Difference of Air Surface temperature using SVR}
  \label{SVR_tas}
\end{figure*}

\begin{figure*}[h]
  \centering
  \textbf{Results with Kernel Ridge Regression}
  \includegraphics[width=\textwidth]{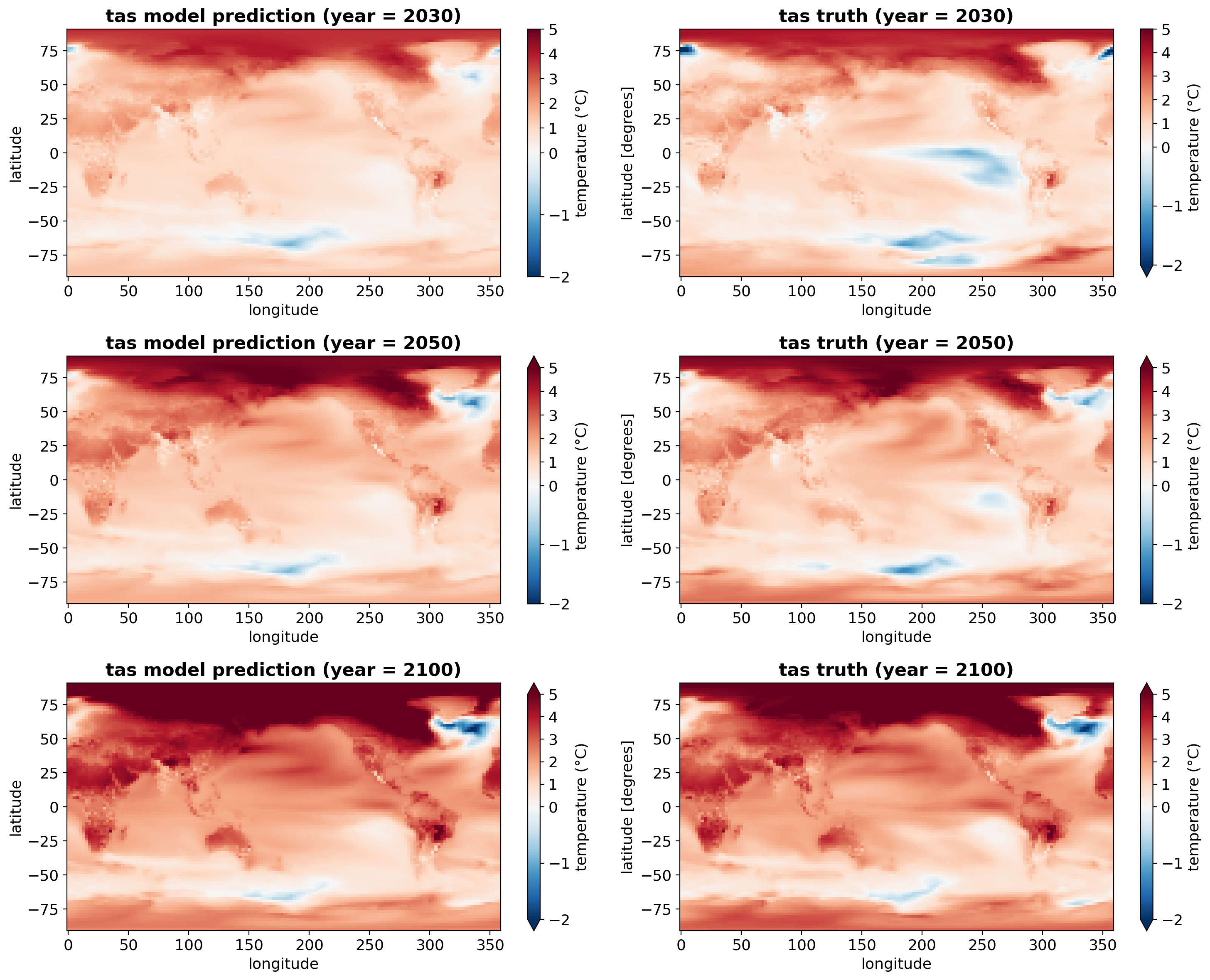}
  \caption{Results of Year Wise Prediction of Air Surface temperature using SVR for different lead times}
  \label{KRR_tas}
\end{figure*}

\begin{figure*}[h]
  \centering
  \includegraphics[width=\textwidth]{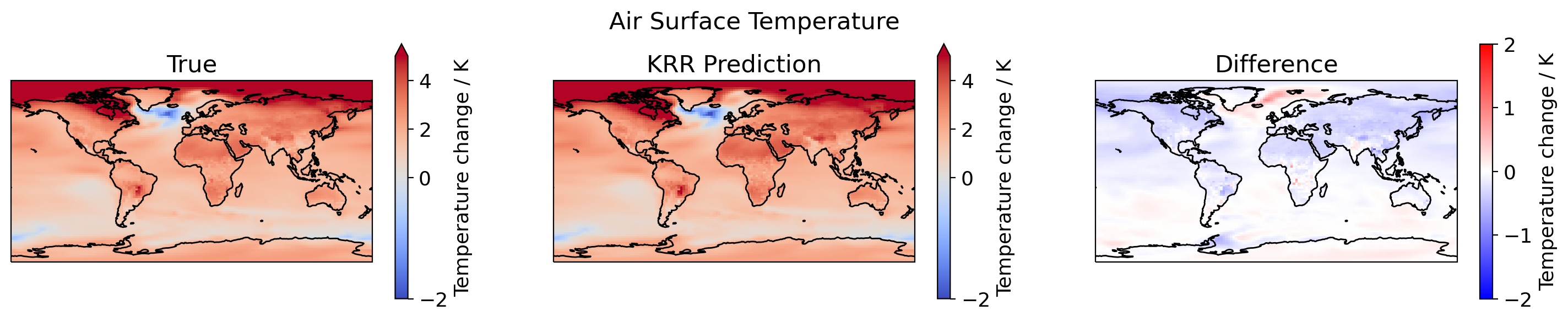}
  \caption{Difference of Air Surface temperature using KRR}
  \label{KRR_tas}
\end{figure*}

\onecolumn

\begin{center}
    \textbf{Visualization using Panoply}
\end{center}

\begin{figure*}[h]

  \centering
  
  \includegraphics[scale = 0.37]{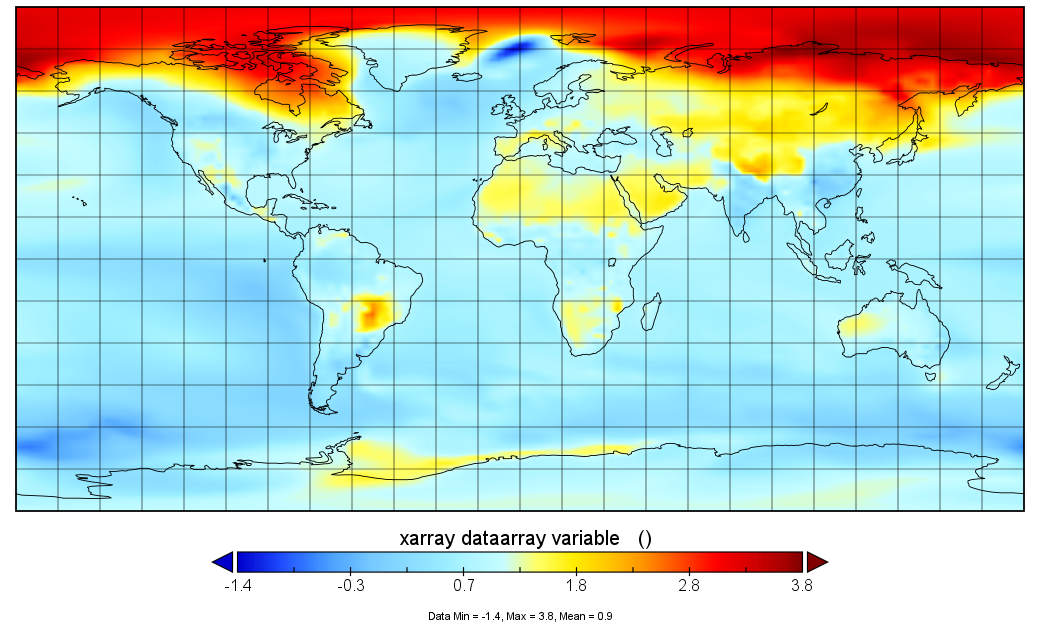}
  \caption{Prediction of GPR }
  \label{GPR_tas}

  \hspace{2 pt}
  \centering
  \includegraphics[scale = 0.37]{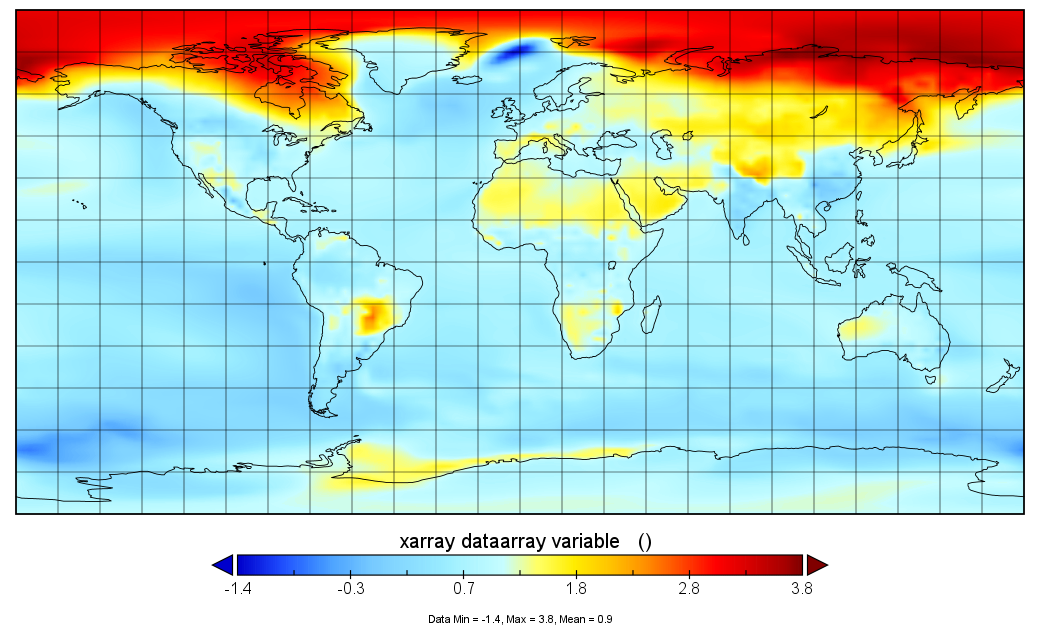}
  \caption{Prediction of Support Vector Regression}
  \label{SVR_tas}
  \hspace{2 pt}
  \centering
  \includegraphics[scale = 0.37]{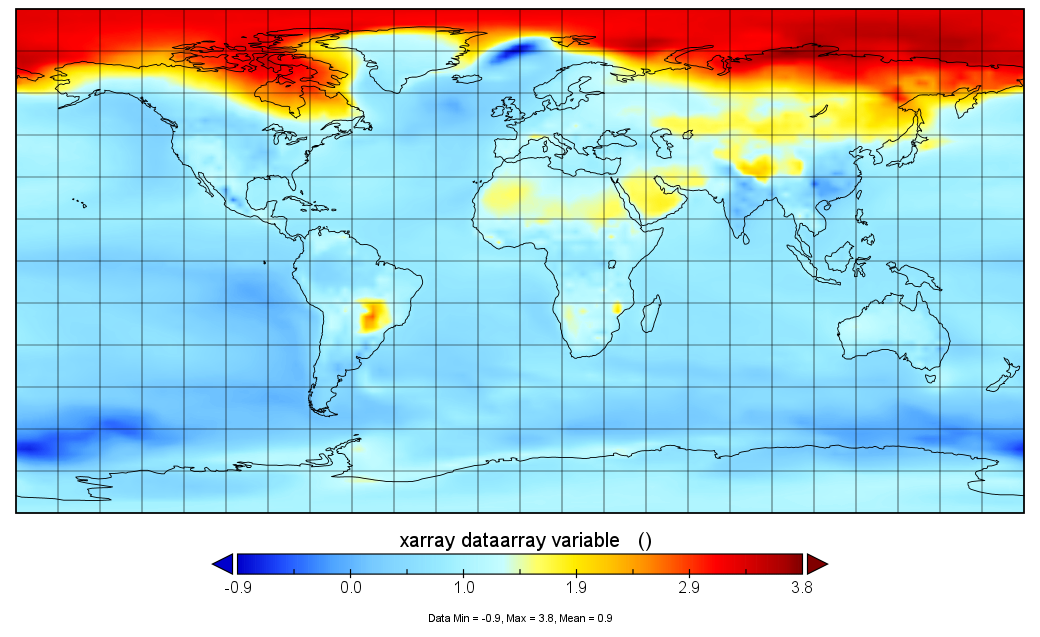}
  \caption{Prediction of Kernel Ridge Regression}
  \label{KRR_tas}

\end{figure*}

\end{document}